\documentclass[10pt,twocolumn,letterpaper]{article}

\usepackage[pagenumbers]{cvpr} 

\usepackage{pifont}

\definecolor{cvprblue}{rgb}{0.21,0.49,0.74}
\usepackage[pagebackref,breaklinks,colorlinks,allcolors=cvprblue]{hyperref}

\def\paperID{*****} 
\def\confName{CVPR}
\def\confYear{2026}

\title{
STARK: Spatio-Temporal Attention for Representation of Keypoints for Continuous Sign Language Recognition
}

\author{Suvajit Patra\\
RKMVERI, Belur\\
{\tt\small suvajit.patra.cs20@gm.rkmvu.ac.in}
\and
Soumitra Samanta\\
RKMVERI, Belur\\
{\tt\small soumitra.samanta@gm.rkmvu.ac.in}
}

\begin{document}
\maketitle
\begin{abstract}
Continuous Sign Language Recognition (CSLR) is a crucial task for understanding the languages of deaf communities. Contemporary keypoint-based approaches typically rely on spatio-temporal encoding, where spatial interactions among keypoints are modeled using Graph Convolutional Networks or attention mechanisms, while temporal dynamics are captured using 1D convolutional networks. However, such designs often introduce a large number of parameters in both the encoder and the decoder. This paper introduces a unified spatio-temporal attention network that computes attention scores both spatially (across keypoints) and temporally (within local windows), and aggregates features to produce a local context-aware spatio-temporal representation. The proposed encoder contains approximately $70-80\%$ fewer parameters than existing state-of-the-art models while achieving comparable performance to keypoint-based methods on the Phoenix-14T dataset.

\end{abstract}    
\section{Introduction}
\label{sec:intro}

Sign Languages (SLs) are natural visual languages, made by and for the deaf communities to exchange information, characterized by the coordinated articulation of arms, facial expressions, and body movements. Translating SL to spoken language is an active area of research due to its significant social impact.

Sign languages consist of both static signs and gestural signs, where static signs consist of a specific pose that represents a gloss, whereas gestural signs consist of a certain sequence of hand, body, and facial articulations that correspond to a gloss. Initial approaches~\cite{BowdenRichardBMVC16, CihanCamgozICCV17, CuiRunpengCVPR17, KollerOscarIJCV18} in Continuous Sign Language Recognition (CSLR) started with 2D Convolutional Neural Network (CNN) based spatial encoders and Hidden Markov Model (HMM) or Recurrent Neural Network (RNN) based temporal decoders. However, these separated spatial–temporal modeling couldn't jointly learn the complex spatio-temporal dynamics.

In keypoints-based CSLR, SignBERT+~\cite{HuHezhenPAMI23} introduces a hand-model-aware self-supervised pre-training framework for sign language understanding that integrates 3D hand mesh reconstruction to capture fine-grained hand gestures and trajectories. MSKA~\cite{GuanMoPR25} decouples keypoints based on body parts and uses multiple streams, where spatial modeling is performed using an attention mechanism on channel-wise subsets and 1D convolution for temporal modeling. Haque \emph{et al.}~\cite{HaqueMdRezwanulICCVW25} used a type of conformer that jointly computes multi-head self-attention spatially and uses 1D CNN for temporal modeling. Min \emph{et al.}~\cite{MinYuecongICCVW25} used a two-stream framework consisting of an RGB stream and a keypoints stream, where a 2D CNN is used in the RGB stream and a Graph Convolutional Network (GCN) is used for the keypoints stream, followed by 1D CNN for temporal modeling and Bi-LSTM for sequence modeling. Also, CoSign~\cite{JiaoPeiqiICCV23} showed a similar approach where spatio-temporal modelling is done using ST-GCN blocks followed by 1D CNN and Bi-LSTM layers for sequence modelling. All these keypoint-based models are based on very high number of learnable parameters both inn encoder and decoder. 

In CSLR, temporal information is dependent on consecutive frames, and there are signer-level speed variability and video-level recording variability. To capture such variability, this paper propose a unified spatio-temporal attention mechanism that adaptively computes correlation scores among intra and inter-keypoints in consecutive frames. We call this Spatio-Temporal Attention for Representation of Keypoints (STARK). This model computes spatial contextual features using an attention mechanism over intra-frame keypoints and computes temporal contextual features using an attention mechanism over inter-frame keypoints in consecutive frames, and finally aggregates them based on attention scores. The proposed encoder shows competitive performance compared to state-of-the-art methods, CoSign~\cite{JiaoPeiqiICCV23} and MSKA~\cite{GuanMoPR25}, with $\approx$ 70\% and $\approx$ 80\% fewer encoder parameters repectively.

\begin{figure*}[ht!]
    \centering
    \includegraphics[width=1\linewidth]{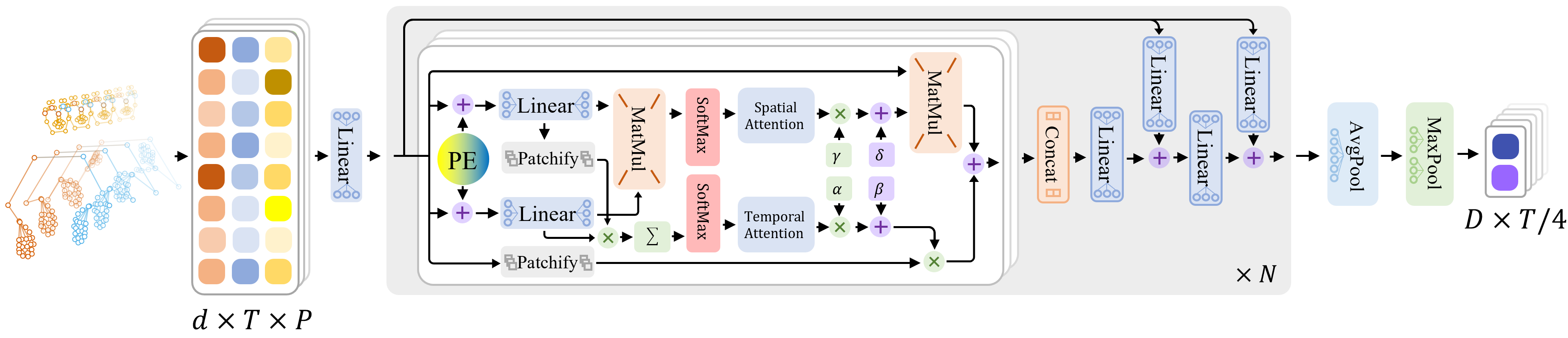}
    \caption{Overview of the proposed STARK model architecture. A sign video is represented as a keypoint tensor $X_{input} \in \mathbb{R}^{d \times T \times P}$ containing $x$, $y$ coordinates and confidence scores for $P$ joints over $T$ frames. The input is projected with a linear layer, followed by the addition of positional encoding. Stacked STARK blocks jointly model temporal relations between the same keypoints across neighboring frames and spatial relations between different keypoints within each frame using spatio-temporal attention. The resulting features are aggregated with average pooling over keypoints and temporally downsampled with max pooling, producing a compact representation of size $D \times T/4$, which is then passed to the gloss decoder for gloss recognition.}
    \label{fig:stark_model}
\end{figure*}

\section{Methodology}
\label{sec:methodology}

Let $I = \{i_1, i_2, \dots, i_T\}$ denote the input sequence of $T$ frames of a sign video, where $i_t \in \mathbb{R}^{P \times d}$ represents the $P$ keypoints with $d$ dimensions ($x$, $y$, and confidence score used for experimental evaluation) at frame $t$. The goal of CSLR is to learn a mapping $f: I \rightarrow J$, where $J = \{j_1, j_2, \dots, j_L\}$ is the corresponding gloss sequence. So, the input sign video is represented as a tensor 
\[
X_{input} \in \mathbb{R}^{d \times T \times P},
\]
Following MSKA~\cite{GuanMoPR25}, we use \emph{four} input streams: body, left (left arm, left hand, and left eye), right (right arm, right hand, and right eye), and face. These are encoded with separate STARK blocks (described below and illustrated in Figure~\ref{fig:stark_model}).

\subsection{Spatio-Temporal Attention for Representation of Keypoints (STARK)}

Given a sign video represented as a sequence of keypoints, the input is denoted as $X_{input} \in \mathbb{R}^{d \times T \times P}$. The goal of the network is to learn a spatio-temporal representation that captures both intra-frame spatial relationships and inter-frame temporal dynamics.

\textbf{Spatio-Temporal Attention Module}  
The core component of STARK is a unified attention module that learns spatial and temporal dependencies jointly. Given input features $X \in \mathbb{R}^{C \times T \times P}$, query and key representations are generated using a linear layer:
\begin{gather*}
    Q, K = FC(X),\\
\end{gather*}
where $Q, K$ are projected in $S$ subspaces with dimensions $C'$, and, as in sign language, recognizing a gloss require aggregation of neighborhood visual information, $K$, and $F$ is converted to $K_{pathches}$, and $X_{pathches}$ using a $pachify$ operation that gives temporally sliding patches based on given parameters: kernel size and stride. Next, the temporal attention is computed over local temporal neighborhoods using :
\[
A_t = \text{softmax}\left(\frac{\sum_{C'}Q \odot K_{pathches}}{C'}\right) \times \alpha + \beta,
\]
where $C'$ denotes the feature dimension, $\odot$ denotes the Hadamard product, and $\alpha, \beta$ are parameters for global temporal attention projection based on different subspaces and keypoints. This attention captures temporal correlations between consecutive frames, where attention scores are calculated for every keypoint and the same keypoint in consecutive neighborhood frames.

The global spatial attention is computed between keypoints within each frame:
\[
A_s = \text{softmax}\left(\sum_{T}\frac{QK^\top}{C'T}\right) \times \gamma + \delta,
\]
where $\gamma, \delta$ are parameters for global spatial attention projection based on different subspaces and keypoints. This models the relationships between different body joints in different subspaces.

The spatial and temporal attention outputs are aggregated to produce the final feature representation:
\[
X_a = \sum_{k}A_t X_{patches} - A_t[k/2] X + A_t[k/2]A_s X,
\]
where $k$ is the kernel size. 

The attention mechanism is followed by a concatenation of attention head outputs, a linear projection layer, along with residual connections and a feed-forward layer:
\begin{align*}
Y &= FC(X_a), \\
Y &= \sigma\left(FC(X) + Y\right), \\
Y &= FFN(Y), \\
X_{out} &= \sigma\left(FC(X) + Y\right),
\end{align*}
where $\sigma$ is the activation function (leaky ReLU used for experimental evaluation).

\textbf{STARK Block}  
Multiple spatio-temporal attention blocks are stacked to progressively learn higher-level representations. Finally the features are aggregated across keypoints using mean/average pooling over keypoints:
\[
H_t = \frac{1}{P} \sum_{p=1}^{P} X^{t,p}_{out}.
\]

As gloss sequences are typically much shorter in length, temporal downsampling is applied using max pooling to obtain compact sequence representations. Let $H \in \mathbb{R}^{T \times D}$ denote the temporal feature sequence, where $T$ is the number of frames and $D$ is the feature dimension. Temporal max pooling is applied along the time dimension to produce

\[
Z = \mathrm{MaxPool}_{t}(H) \in \mathbb{R}^{T' \times D},
\]

where $T' < T$. The resulting representation $Z$ encodes the spatio-temporal dynamics of the sign sequence and is used for subsequent sequence modeling and recognition.

\subsection{Decoder}

The outputs (body, left, right, face) from the STARK encoders are concatenated in the following stream for decoding: fuse (body, left, right, face), left (left, face), right (right, face), and body (body). These streams are decoded to glosses using a decoder inspired by MSKA~\cite{GuanMoPR25}, which comprises a linear projection, temporal positional encoding, batch normalization, and a feedforward layer with a residual connection.

\subsection{Loss Function}
We train the model using Connectionist Temporal Classification (CTC) loss following previous approaches~\cite{AlyamiSarahTMM26, AlyamiSarahICCVW25, CamgozNecatiCihanCVPR18, BowdenRichardBMVC16, HuHezhenPAMI23}. Additionally, following MSKA~\cite{GuanMoPR25}, we use cross-distillation loss using the Kullback-Leibler (KL) divergence using ensemble gloss probabilities.

\section{Experiments \& Results}
\label{sec:exp-results}

\begin{table*}[t]
\centering
\small
\caption{Comparison with previous approaches on Phoenix-2014T dataset. $\downarrow$ denotes lower is better.}
\begin{tabular}{l c c c cc}
\toprule
\textbf{Method} & \textbf{Pre-Training} & \textbf{Encoder Params}$\downarrow$ & \textbf{Decoder Params}$\downarrow$ & \multicolumn{2}{c}{\textbf{Phoenix-2014T}} \\
 &  & & & Dev$\downarrow$ & Test$\downarrow$ \\
\midrule

TwoStream-SLR \cite{ChenYutongNEURIPS22} & \checkmark & - & - & 27.1 & 27.2 \\
SignBERT+ \cite{HuHezhenPAMI23} & \checkmark & - & - & 32.9 & 33.6 \\
CoSign \cite{JiaoPeiqiICCV23} & \ding{55} & $\approx$ 10M & $\approx$ \textbf{18M} & \textbf{19.5} & \textbf{20.1} \\
MSKA \cite{GuanMoPR25} & \ding{55} &  $\approx$ 15M &  $\approx$ 28M & 20.1 & 20.5 \\
Ours & \ding{55} &  $\approx$ \textbf{3M} &  $\approx$ 28M & 21.0 & 21.9 \\

\bottomrule
\end{tabular}
\label{tab:results}
\end{table*}

\subsection{Datasets}

\textbf{Phoenix14T} 
The RWTH-PHOENIX-Weather 2014T (Phoenix14T)~\cite{CamgozNecatiCihanCVPR18} dataset is one of the most widely used benchmarks for continuous sign language recognition and translation. It consists of weather forecast recordings from the German public television channel PHOENIX, featuring 9 signers, performing German Sign Language (DGS). The dataset provides sign-language videos, along with their corresponding German glosses and text transcripts, for sign-language recognition and translation research. Phoenix14T contains 8,257 annotated video sequences over a vocabulary of 1066 sign glosses and follows a split of 7,096 training samples, 519 validation samples, and 642 test samples. The videos exhibit real-world challenges such as multi-signer motion variability, and varying signing speeds.

\subsection{Data Pre-Processing \& Augmentations}
We use HRNet~\cite{WangJingdongTPAMI20} keypoints data from the Phoenix14T dataset~\cite{CamgozNecatiCihanCVPR18}, containing 133 keypoints for each frame. Out of these 133, we select 79 keypoints (including body, left hand, right hand, and face keypoints) following the previous approach~\cite{GuanMoPR25}. We use $x$ and $y$ coordinates, along with the confidence scores. The coordinates are in the pixel space with the origin at the top-left corner of the image, and the confidence scores range from 0 to 1. Following prior approaches~\cite{SuvajitPatraPAA25, GuanMoPR25}, we augment the keypoints data in the following ways: 1) The coordinates are normalized in -1 to 1, 2) To incorporate signing speed variability, we downsample and upsample the input keypoints sequence between $\times0.5$, and $\times1.5$ randomly, 3) The keypoints are randomly rotated in the 2D frame within the range of $[-15^\circ, 15^\circ]$.

\subsection{Experimental Settings}
The model is implemented using PyTorch\footnote{\url{https://pytorch.org/}} library in Python 3.10\footnote{\url{https://www.python.org/}}. The STARK block begins with an input projection layer that projects the input from 3 channels to 64 channels, followed by \emph{four} spatio-temporal attention modules with output channel dimensions of 64, 96, 128, and 256, each using 6 attention heads. 

The training settings for CSLR are as follows: Adam optimizer (weight decay = $1e^{-3}$), Cosine Annealing scheduler ($T_{max}=100$), an initial learning rate of $1e^{-3}$, and a batch size of 8. For the CTC decoder, greedy search is used during training with a beam width of 1, while during inference, beam search is applied with a beam width of 5. The training is executed on an Ubuntu server with Intel Gold processors, along with an NVIDIA A100 GPU.

Following contemporary approaches~\cite{KollerOscarCVIP15, AlyamiSarahICCVW25, AlyamiSarahTMM26, MinYuecongICCVW25, HaqueMdRezwanulICCVW25, LiZechengARXIV25, GuanMoPR25}, we use Word Error Rate (WER) as the evaluation metric for CSLR.

\subsection{Comparison with state-of-the-art methods}

In the CSLR task, several keypoint-based methods have been proposed~\cite{ChenYutongNEURIPS22, HuHezhenPAMI23, JiaoPeiqiICCV23, GuanMoPR25} as tabulated in Table~\ref{tab:results}. Some approaches leverage pre-training on large-scale datasets before fine-tuning for the downstream CSLR task. In contrast, without pre-training, our method achieves 21.0 WER on the validation set and 21.9 WER on the test set of the Phoenix-14T dataset. Compared to the state-of-the-art keypoints-based method CoSign~\cite{JiaoPeiqiICCV23}, our method differs by 1.5 and 1.8 WER on the Phoenix-14T validation and test sets, respectively. However, our encoder contains approximately 3 million parameters, which is about 70\% fewer than the 10 million parameters used in CoSign. Additionally, compared to MSKA~\cite{GuanMoPR25}, our method differs by 0.9 and 1.4 WER on the Phoenix-14T validation and test sets, respectively, while using about 80\% fewer encoder parameters (3 million vs. 15 million).
\section{Conclusion}
\label{sec:conclusion}

In this work, we proposed a unified spatio-temporal attention encoder for keypoint-based Continuous Sign Language Recognition. Unlike conventional approaches that separately model spatial and temporal relationships, our model jointly captures spatial interactions among keypoints and temporal dependencies within local windows through a unified attention mechanism. The encoder learns local context-aware spatio-temporal representations efficiently, requiring 70–80\% fewer parameters than existing state-of-the-art keypoint-based encoders, while maintaining competitive performance on the Phoenix-14T dataset. Although the results are promising, further study is required to fully explore the effectiveness of the proposed method.
{
    \small
    \bibliographystyle{ieeenat_fullname}
    \bibliography{main}
}


\end{document}